\pdfoutput=1

\documentclass[11pt]{article}

\usepackage[preprint]{acl}

\usepackage{times}
\usepackage{latexsym}

\usepackage[T1]{fontenc}

\usepackage[utf8]{inputenc}

\usepackage{microtype}
\usepackage{mdframed}

\usepackage{inconsolata}
\NewDocumentCommand{\zj}{mO{}}{\textcolor{blue}
{\textsuperscript{\textit{zj}}\textsf{\textbf{\small[#1]}}}}
\newcommand{\vpara}[1]{\vspace{0.05in}\noindent \textbf{#1 }}
\newcommand{\tabincell}[2]{\begin{tabular}{@{}#1@{}}#2\end{tabular}}
\usepackage{xcolor}

\definecolor{darkgreen}{RGB}{0,100,0}

\usepackage{appendix}
\usepackage{titletoc}

\usepackage{url}
\usepackage{subfigure}
\usepackage{multirow}
\usepackage{enumitem}
\usepackage{stmaryrd}
\usepackage{array}
\usepackage{threeparttable}
\usepackage{blindtext}
\usepackage{amssymb}

\usepackage[ruled,vlined,linesnumbered,longend]{algorithm2e}
\usepackage{svg}
\usepackage{placeins}

\usepackage{soul}
\usepackage[utf8]{inputenc}
\usepackage{graphicx}
\usepackage{amsmath}
\usepackage{booktabs}
\usepackage{lipsum}
\usepackage{listings}
\usepackage{xcolor, soul}
\sethlcolor{blue!10}
\usepackage{colortbl}
\usepackage{bbm}
\newcommand{\model}{\textsc{PiNose}\xspace}
\newcommand{\smodel}{\textsc{PiNose} }

\definecolor{sgreen}{HTML}{F3FADF} 
\definecolor{mgreen}{HTML}{E0EAB5} 
\definecolor{dgreen}{HTML}{CDDC8C} 
\definecolor{ddgreen}{HTML}{B8CF61} 

\definecolor{'sred'}{HTML}{FFEAE8}

\usepackage{amsmath,amsfonts,bm}

\def\eqref#1{equation~\ref{#1}}

\def\1{\bm{1}}

\def\rvb{{\mathbf{b}}}

\def\rvq{{\mathbf{q}}}
\def\rvr{{\mathbf{r}}}

\def\rmH{{\mathbf{H}}}

\def\rmW{{\mathbf{W}}}

\DeclareMathAlphabet{\mathsfit}{\encodingdefault}{\sfdefault}{m}{sl}
\SetMathAlphabet{\mathsfit}{bold}{\encodingdefault}{\sfdefault}{bx}{n}

\title{
Transferable and Efficient Non-Factual Content Detection \\
via Probe Training with Offline Consistency Checking
}

\author{
    Xiaokang Zhang \textsuperscript{1\footnotemark[2]\footnotemark[3]},
    Zijun Yao\textsuperscript{2\footnotemark[2]}, 
    Jing Zhang\textsuperscript{1\footnotemark[1]}, \\
\textbf{Kaifeng Yun,}\textsuperscript{2} \textbf{Jifan Yu,}\textsuperscript{2} \textbf{Juanzi Li,}\textsuperscript{2} \textbf{Jie Tang}\textsuperscript{2} \\
\textsuperscript{1}School of Information, Renmin University of China, Beijing, China, \\
\textsuperscript{2}Department of Computer Science and Technology, Tsinghua University,  Beijing, China \\
\texttt{\{zhang2718,zhang-jing\}@ruc.edu.cn},\\ 
\texttt{\{yaozj20, ykf21,yujf21\}@mails.tsinghua.edu.cn},\\ 
\texttt{\{juanzi,jietang\}@tsinghua.edu.cn}
}

\begin{document}

\maketitle
\renewcommand{\thefootnote}{\fnsymbol{footnote}}
\footnotetext[2]{Equal Contribution.}
\footnotetext[3]{Work was done when interned at Zhipu AI.}
\footnotetext[1]{Corresponding author.}
\renewcommand*{\thefootnote}{\arabic{footnote}}

\begin{abstract}
Detecting non-factual content is a long-standing goal to increase the trustworthiness of large language models (LLMs) generations.
Current factuality probes, trained using human-annotated labels, exhibit limited transferability to out-of-distribution content,
while online self-consistency checking imposes extensive computation burden due to the necessity of generating multiple outputs.
This paper proposes \model, which trains a probing model on offline self-consistency checking results, thereby circumventing the need for human-annotated data and achieving transferability across diverse data distributions. As the consistency check process is offline, \model reduces the computational burden of generating multiple responses by online consistency verification. 
Additionally, it examines various aspects of internal states prior to response decoding, contributing to more effective detection of factual inaccuracies. 
Experiment results on both factuality detection
and question answering benchmarks  show that \model achieves surpassing results than existing factuality detection methods.
Our code and datasets are publicly available on this \href{https://github.com/Pinocchio42/PiNose}{anonymized repository}.
\end{abstract}

\section{Introduction}
\label{sec:intro}

Large language models (LLMs), after pre-training on massive corpora~\cite{gpt3,llama,mistral}, show a surprising ability to generate knowledgeable content~\cite{sun2022recitation,yu2022generate}.
Although this ability facilitates a wide range of applications, such as question answering (QA)~\cite{abdallah2023generator,liu2021generated,li2022self} and information retrieval~\cite{mao2020generation,ma2023query}, 
the propensity of LLMs to occasionally produce non-factual knowledge~\cite{truthfulqa,wang2023survey} potentially hinders the practical utilization of generated content.
Thus, it is necessary \textit{to detect whether LLMs generate non-factual content}.

Previous studies offer evidence that the internal representation vectors in LLMs determine whether they produce factual answers to the input question~\cite{azaria2023internal,kadavath2022language,zou2023representation}.
Specifically, the factual behavior entailed is extracted from the feed-forward layer activations of tokens before the generated content using linear probes~\cite{alain2016understanding,belinkov2022probing}.
However, their construction relies on the labor-intensive process of annotating natural language questions, as well as labeling LLMs' outputs with factuality annotations, a factor that limits their applicability to questions and responses with unseen distributions.

To avoid the annotation process, the most recent studies detect non-factual content via online self-consistency checking~\cite{wang2022self}.
They assume that if LLMs give contradictory responses to the same prompt, the model is more likely to hallucinate to give that answer~\cite{elazar2021measuring}. In this way, detecting non-factual content is reduced to the mutual-entailment analysis among multiple generations, which is usually realized as natural language inference (NLI) models~\cite{kuhn2023semantic,manakul2023selfcheckgpt} or heuristic comparison of the hidden representation similarity~\cite{anonymous2023inside}. However, self-consistency checking introduces extensive computation overhead to sample multiple responses. In addition, due to the lack of training process, these methods are less robust than previous factuality probes.

Giving these limitations of existing methods, we propose \model, a method to
\underline{p}red\underline{i}ct \underline{no}n-factual respon\underline{se}s from LLMs.
The main idea of \model is to construct a probing model that learns from offline self-consistency checking.
It aims to present two core advantages over existing methods:

\textbf{Transferability.} 
Comparing with existing probing methods, \model eliminates human annotation for training data.
This is achieved with bootstrapped natural language questions and generated pseudo factuality labels through an \textit{offline} consistency checking mechanism.
Moreover, as \model does not rely on specific training data, it transfers effortlessly to any different data distributions.

\textbf{Efficiency and Effectiveness.}
Comparing with online consistency checking, \model avoids the computational burden associated with multiple generations during inference, thus enhancing time efficiency.
Additionally, by analyzing the continuous internal representations of LLMs  
rather than discrete tokens in the response,
\model gains access to a broader spectrum of information, enhancing its prediction effectiveness.

We conduct comprehensive experiments on established factuality detection benchmarks and variations of QA datasets. Our results reveal several key findings: (1) \smodel outperforms supervised probing-based baselines by 7.7-14.6 AUC across QA datasets, despite being trained without annotated labels. (2) Moreover, our \smodel achieves significant performance improvements (3-7 AUC) compared to unsupervised consistency checking baselines, while also demonstrating superior time efficiency. (3) Additionally, the dataset generated via offline self-consistency checking shows promise for transferring to probe various LLMs.

\section{Preliminaries}
\label{sec:relate}

This study concentrates on identifying non-factual content using decoder-only LLMs. It begins by formally defining the task and then elaborates on decoder-only LLMs and the construction of probes for these models. Additionally, it discusses the distinctions between online and offline self-consistency checking.

\paragraph{Task Definition.}
Formally, given a question $\rvq=\langle q_1, q_2, \dots, q_n \rangle$ composed of $n$ tokens, and its corresponding response $\rvr=\langle r_1, r_2, \dots, r_m \rangle$ consisting of $m$ tokens, non-factual content detection aims to assign a binary judgment $f\in\{\mathtt{True}, \mathtt{False}\}$, determining the factual correctness of $\rvr$.
For example, given the question ``\textit{What is the capital city of China?}'', ``\textit{Beijing}'' serves as the response with $\mathtt{True}$ judgement, while ``\textit{Shanghai}'' is classified as $\mathtt{False}$.
Without losing generality, we allow $\rvq$ to be null (\textit{i.e.,} $\rvq=\varnothing$), in which scenario, the task evaluates the factuality of the standalone assertion.

To detect whether LLMs generate non-factual content, in our setting, the response $\rvr$ is usually sampled from an LLM. In this case, we expect the model to assess the factuality of content generated by itself without the need of another LLM. 

\paragraph{Decoder-only LLMs.}
Decoder-only LLMs comprise a stack of Transformer decoder layers~\cite{vaswani2017attention}. 
After embedding tokens into the hidden representation $\rmH^{(0)}$, each layer manipulates the hidden representation of the previous layer as follows\footnote{We omit residual connection~\cite{he2016deep} and layer normalization~\cite{ba2016layer} for simplicity.}:
\begin{equation*}\small
\rmH^{(l)} = \mathtt{FFN}\left(\mathtt{Attn}\left(\rmH^{(l-1)}\right)\right),
\end{equation*}\normalsize
where $\mathtt{Attn}(\cdot)$ is the attention mechanism.
$\mathtt{FFN}(\cdot)$ is the feed-forward network composed of two consecutive affine transformations and activation functions.
Thus, the intermediate representations of decoder-only LLMs are usually extracted from the output of $\mathtt{FFN}(\cdot)$ operation.
In the following of this paper, we denote the hidden representation of the $i^\text{th}$ token extracted from layer $l$ as $\rmH^{(l)}[i]$.

\paragraph{Language Model Probing.}
Probing method extracts implicit information from the intermediate representation.
It is usually implemented as a simple classification model that maps from the hidden representation of certain token into discrete classes.

The probe in \model is a two-layer feed-forward network with binary classification outputs:

\vspace{-0.1in}\footnotesize
\begin{equation}
\label{equ:probe}
    \mathtt{Probe}\left(\rmH^{(l)}[i]\right) = \sigma_2\left(\rmW_2\sigma_1\left(\rmW_1\rmH^{(l)}[i] + \rvb_1\right) + \rvb_2\right),
\end{equation}\normalsize

\noindent where $\sigma_1$ is the $\texttt{Sigmoid}$ function, $\sigma_2$ is any non-linear function, and $\rmW, \rvb$ are trainable parameters.
$\mathtt{Probe}\left(\rmH^{(l)}[i]\right)$ is the probability for $\mathtt{True}$.

\paragraph{Consistency Checking.}
Consistency checking requires LLMs to generate multiple responses towards the same question, and utilize these semantic consistency to judge whether the generations are correct.
Previous methods for non-factual content detection are \textbf{\textit{online}} self-consistency checking, where LLMs need to generate extensive responses to answer a single question to obtain factuality labels.
Our method falls into the \textit{\textbf{offline}} consistency checking category, where consistency checking is solely used to generate labels for training the probe. During online checking, LLMs only need to produce a single response and obtain the factuality label from the probe.
\section{Methodology}
\label{sec:method}

\begin{figure*}[!ht]
    \centering
    \includegraphics[width=0.97\linewidth]{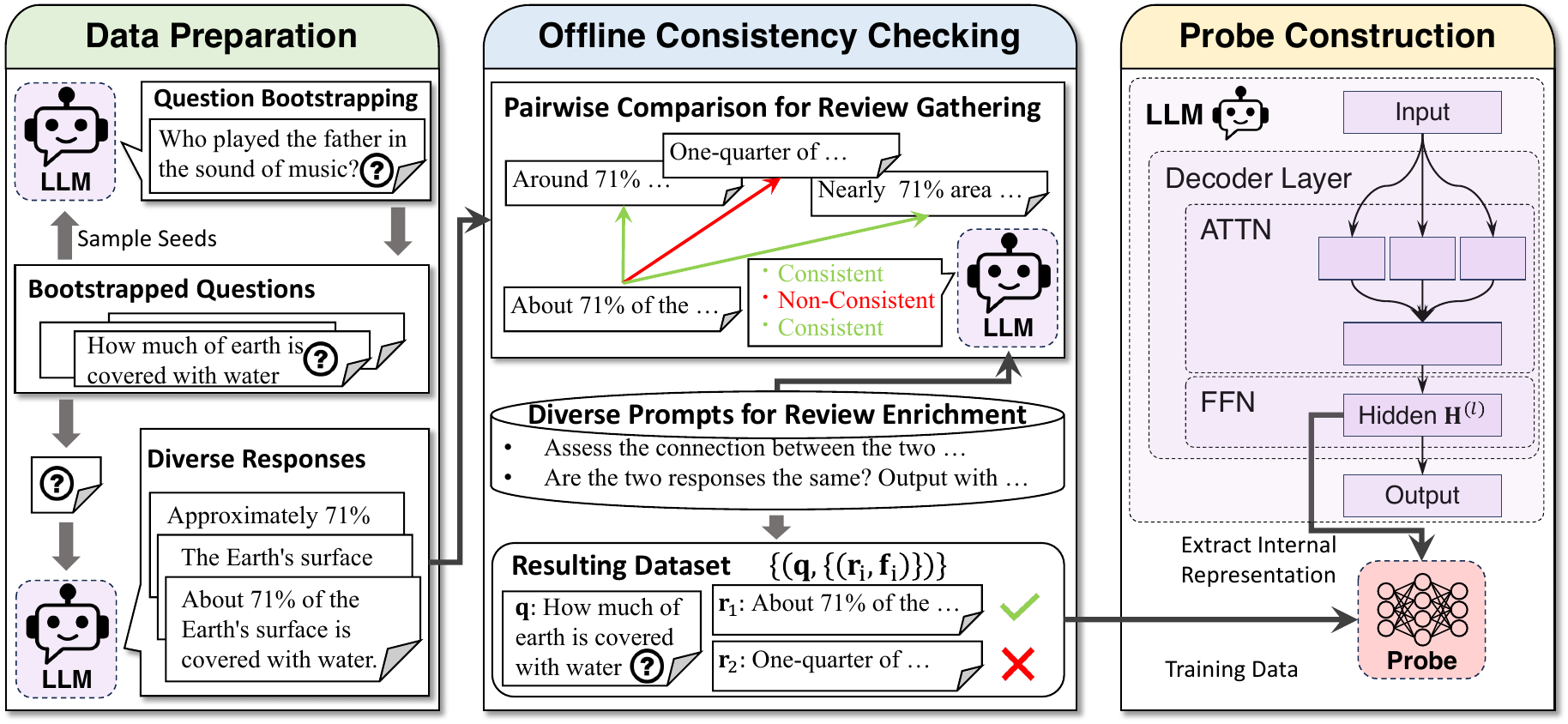}
    \caption{The overall architecture of \model.
    \label{fig:framework}
    }
\end{figure*}

The construction of \model involves three main stages:
(1) In the data preparation stage, we bootstrap natural language questions and generate multiple responses, which together serve as model inputs;
(2) In the offline consistency checking stage, we employ a peer review mechanism to generate pseudo factuality label for each response;
(3) In the probe construction stage, these pseudo factuality labels are used to train a language model probe.
Figure~\ref{fig:framework} illustrates the overall process.

\subsection{Stage 1: Data Preparation}

In alignment with the requirements of the non-factual content detection task, the supporting data consists of three elements: natural language questions $\rvq$, their corresponding responses $\rvr$, and factuality labels $f$.
This stage concentrates on generating large-scale data containing the initial two elements (\textit{i.e.,} $\rvq$ and $\rvr$).

\textbf{Question Bootstrapping.}
\model leverages natural language questions to prompt LLMs to generate responses for consistency checking.
However, natural language questions are not always available across all domains.
Furthermore, both the diversity and quantity of questions can significantly impact the quality of the prepared data. Therefore, we aim to enable LLMs to bootstrap questions with minimal human involvement.

Fortunately, as \citet{honovich2022unnatural} point out, high-performing language models show significant capacity in question generation.
Inspired by these findings, we manually annotate a set of seed questions and employ them as demonstrations for LLMs to generate a large volume of questions via in-context learning~\citep[ICL]{gpt3}. 
To enhance diversity in generation, we broaden the scope of seed questions by incorporating the generated ones and sample diverse combinations from the seed questions for subsequent generation. 
Detailed prompt for question generation is provided in Figure~\ref{prompt:questiongeneration} of Appendix~\ref{sec:prompt}.

\textbf{Diverse Response Generation.}
We use previously generated questions as input for LLMs to generate multiple responses for subsequent consistency checking.
We design two strategies to encourage the diversity of multiple responses to the same input question.
(1) From the perspective of decoding, we adjust the decoding strategy by applying a greedy sampling method with a relatively high sampling temperature ($t=1$).
(2) From the perspective of model input, we instruct LLMs to answer a question using a variety of prompts (as shown in Figure~\ref{prompt:responsegeneration} in Appendix~\ref{sec:prompt}).

The outcome of this stage is a dataset containing questions paired with multiple responses, designated as $\{\left(\rvq, \{\rvr_i\}\right)\}$, where the number of responses $k = |\{\rvr_i\}|$ serves as a hyperparameter that determines the quantity of responses per question for the subsequent consistency check.

\subsection{Stage 2: Offline Consistency Checking}

We engage LLMs in the offline consistency check process via a peer review mechanism.
First, we gather reviews by asking LLMs to determine for each response whether it is consistent with other responses.
Then, we enrich reviews by sampling multiple consistency judgements by varying model inputs.
Finally, we integrate reviews to form the pseudo-factuality label for each response and filter out low-quality responses.

\textbf{Review Gathering.}
Formally, consistency reviewing involves asking an LLM to evaluate whether the response $\rvr_i$ to the question $\rvq$ is semantically consistent with other responses (\textit{i.e.,} $\rvr_j, j\neq i$). 
If $\rvr_i$ has equivalent meaning with other responses, it is considered factual. 
To ensure unambiguous judgment, we require the LLM to make pairwise comparisons with other $k-1$ responses. For each comparison, it must output one of three labels: ``\texttt{Consistent}'', ``\texttt{Neutral}'', or ``\texttt{Non-Consistent}''. 
To achieve this, we specify the output format with in-context demonstrations and prompt instructions (as shown in Figure~\ref{prompt:reviewgeneration} in Appendix~\ref{sec:prompt}).

\textbf{Review Enrichment.}
To enhance the diversity of reviews, we introduce variability in the input provided to the LLM during consistency assessments. Recognizing the significant impact of demonstrations on LLM judgments in ICL~\cite{wang2023label}, we utilize a range of diverse demonstration combinations for ICL to elicit varied reviews from the LLM for each pairwise comparison. Diverse demonstrations facilitate the collection of multiple reviews, each potentially providing a unique perspective. In total, we gather $N$ round of reviews for each pairwise comparison, where $N$ is a hyper-parameter.

\textbf{Integration and Filtering.}
We integrate $N$ reviews for each pairwise comparison, and subsequently integrate $k-1$ pairwise comparisons for each response through the same majority voting mechanism. Here is how the voting works: we first consider \texttt{Neutral} consistency judgement as an abstention for voting.
Then, to guarantee the quality of the final dataset, we exclude controversial judgements where no single label (\texttt{Consistent}, \texttt{Neutral}, \texttt{Non-Consistent}) receives over $50\%$ of the votes. 
This step ensures that only the most widely agreed-upon judgements are retained for analysis. Finally, we assign the factuality label $\mathtt{True}$ ($\mathtt{False}$) to responses that are predominantly considered consistent (non-consistent) with others.

This stage outputs the dataset with full elements for consistency checking, \textit{i.e.,} $\{(\rvq, \{(\rvr_i, f_i)\})\}$.

\subsection{Stage 3: Probe Construction}

\model predicts the factuality of responses via a probing model, as defined in Equation~\ref{equ:probe}.
To be more specific, \smodel integrates the response with the question, both formatted according to the template outlined in Figure~\ref{prompt:probeconstruction} in Appendix~\ref{sec:prompt}, into the LLM for detection. Subsequently, the hidden representation of the last token in the response at the middle layer of the LLM is employed as the input for the probing model.
We train the probing model to maximize the probability of the factuality label while freezing all the parameters of the LLM.
Formally, the construction process of the probe optimizes the following cross-entropy loss:

\begin{equation}\notag\small
     \text{loss} = -\sum_{\{\rvq,\rvr_j,f_j\}}\text{log}\, \texttt{Probe}\Big(\rmH^{(l)}[i]\Big) \mathbbm{1}\Big(f_j=\mathtt{True}\Big),
\end{equation}\normalsize

\noindent where $\mathbbm{1}(\cdot)$ is the indicator function, $i$ is the index of the last input token, and $l$ represents half the layer number of the LLM to be detected.

\subsection{Discussion}

We discuss the rationality of \model and the involvement of LLMs for implementing \model.

The reason of why \model successfully detect non-factual responses comes from the model and data perspective.
(1) \textbf{Model property.} 
LLMs are well-calibrated after massive pre-training~\cite{kadavath2022language,zhu2023calibration}.
This indicates that for non-factual responses, LLMs tend to assign less probability, while preserve relatively high probability for factual generations.
This calibration property guarantees the feasibility of distilling factuality detection dataset from offline consistency checking.
It also suggests that the internal states of LLMs tracks whether they are producing factual contents, which \model tries to uncover with probing model.
(2) \textbf{Data quality.} Offline consistency checking gathers diverse instances of inconsistency between responses from LLMs, potentially enhancing the quality of training data for the probing model. Consequently, it enables the model to address a broader range of inconsistency scenarios compared to online consistency checking.
Moreover, as the data collection process is fully automated, the dataset can be significantly larger than existing training data for factuality probes. 
The feasibility of this principle is also widely verified in distant supervision~\cite{quirk2016distant}.

To implement \model, LLMs are multiply invoked during the construction process, including 
data preparation, peer reviewing in consistency checking, and finally non-factual detection. For a coherent implementation, we employ the same LLM for detecting the factuality of responses as the one used for generation and checking consistency. This implementation strategy aligns with our setting, where no third-party LLM is available, and it also enhances the transferability of our method.

\section{Experiment}
\label{sec:exp}

We conduct experiments to examine the performance of \model by comparing it to baseline methods for factuality detection.
Additionally, we assess its transferability and efficiency.

\subsection{Experiment Setup}

\subsubsection{Datasets}
The datasets include both factuality detection benchmark and variations of QA datasets.
We introduce the purpose to incorporate each dataset and their data specifications.
Detailed statistics are shown in Table~\ref{tb:datastatistics}.

\textbf{Benchmark.}
We follow previous research to use \textbf{True-False} benchmark~\cite{azaria-mitchell-2023-internal}.
True-False provides statements generated by LLMs along with corresponding factuality labels examined by humans. It does not include questions for each statement (\textit{i.e.,} $\rvq=\varnothing$).
True-False comes with both training dataset and test dataset.

\textbf{Variation of QA.}
Given that probing statements, such as those in True-False dataset, is less practical compared to examining responses to questions from LLMs, for practical evaluation, we establish test sets based on existing QA datasets: \textbf{Natural Questions (NQ)}~\cite{naturalquestion}, \textbf{TriviaQA}~\cite{triviaqa}, and \textbf{WebQ}~\cite{webq}.
In particular, we sample $1,000$ questions from each dataset and employ Llama2-7B~\cite{llama2} to generate responses accordingly.
The factuality label for each response is annotated by human annotators through comparing with the original ground-truth of each question.

\begin{table}[!t]
    \centering
    \scalebox{0.86}{
    \begin{tabular}{lccccc}
        \toprule
         & True-False & NQ & TriviaQA & WebQ \\
        \midrule
        \#Train & $5,000$ & $\mathtt{N/A}$ & $\mathtt{N/A}$ & $\mathtt{N/A}$ \\
        \#Test & $1,000$ & $1,000$ & $1,000$ & $1,000$ \\
        \%$\mathtt{True}$ & $40.5$ & $46.6$ & $49.5$ & $58.3$\\
        \bottomrule
    \end{tabular}
    }
    \caption{Data statistics. 
    \#Train and \#Test are the number of instances in the training data and test set, respectively.
    \%$\mathtt{True}$ is the ratio of $\mathtt{True}$ labels in the test set.}
    \label{tb:datastatistics}
\end{table}

\subsubsection{Baselines}

We compare \model against probing-based and consistency-checking-based methods. Additionally, to ensure a comprehensive comparison, we implement heuristic confidence-based methods as baselines.

\begin{itemize}[leftmargin=*,itemsep=0pt,parsep=0pt]
\item \textbf{Probing Based:}
\textbf{SAPLMA}~\cite{azaria2023internal} utilizes a feed-forward neural network for factuality detection, trained on the True-False training data. 
\textbf{RepE}~\cite{zou2023representation} conduct principal component analysis (PCA) on the internal representations of True-False training data.
It selects a factual direction vector using the factuality labels. 
During testing, RepE compute the dot product between the internal representation of the given response and the factual direction vector.

\item \textbf{Consistency Checking Based:}
We compare against SelfCheckGPT~\cite{selfcheckgpt}, which performs factuality detection via online self-consistency checking.
We implement two variants.
\textbf{SelfCheckGPT-NLI} (SCGPT-NLI) uses a BERT-based NLI model~\cite{williams-etal-2018-broad} for consistency checking, while \textbf{SelfCheckGPT-Prompt} (SCGPT-PT) elicits LLM itself to evaluate the consistency between two responses via prompt.

The prompt used for SCGPT-PT is shown in Figure~\ref{prompt:selfcheckGPT} in the Appendix.

\item \textbf{Confidence Based:}
We also utilize model confidence as an indicator for factuality detection.
\textbf{Perplexity-AVE} (PPL-AVE) and \textbf{Perplexity-Max} (PPL-MAX)~\cite{kadavath2022language,azaria2023internal,zou2023representation} quantify the average and maximum token-level probabilities of statements within each test set generated by the evaluated LLM. 
\textbf{It-is-True}~\cite{azaria2023internal} compares the probabilities between sentences ``\textit{It is true that} $\rvq||\rvr$.'' and ``\textit{It is false that} $\rvq||\rvr$.'', where $||$ denotes concatenation.

\end{itemize}

It is worth noting that probing based baselines rely on training data.
We thus implement them using the training dataset from True-False.
Besides, as SCGPT-PT and SCGPT-NLI needs input questions to generate multiple responses, it is infeasible to test on True-False, where we mark their results as ``N/A'' in Table~\ref{tab:overall}. 

\subsubsection{Evaluation Metrics}
We follow conventions~\cite{azaria-mitchell-2023-internal} in factuality detection, employing the area under the receiver operating characteristic curve (AUC) and accuracy (ACC) as evaluation metrics.

\subsubsection{Implementation Details}

To implement \model, we uniformly use Llama2-7B for data preparation, consistency checking, and factuality detection.
For hyperparameters, we set the number of sampled responses $N$ to $9$ and the round of peer review $k$ to $7$.
For fair comparison, we also allow SelfCheckGPT to generate $N=9$ responses for consistency checking.
The training dataset for \smodel consists of $20,000$ constructed triplets $\{(\rvq, \rvr_i, f_i)\}$. The threshold for calculating accuracy is determined by selecting the value that yields the highest accuracy among $100$ validation instances partitioned from the test sets.

\begin{table}[!t]
    \centering
    \scalebox{0.79}{
    \setlength{\tabcolsep}{2pt}
    \begin{tabular}{@{}l|@{ }cc@{ }|@{ }cc@{ }|@{ }cc@{ }|@{ }cc@{}}
        \toprule
         & \multicolumn{2}{c|@{ }}{True-False} & \multicolumn{2}{c|@{ }}{NQ} & \multicolumn{2}{c|@{ }}{TriviaQA} & \multicolumn{2}{c}{WebQ} \\
        \cmidrule(r){2-3}\cmidrule(r){4-5}\cmidrule(r){6-7}\cmidrule(r){8-9}  
        & AUC & ACC & AUC & ACC & AUC & ACC & AUC & ACC \\
        \midrule
        RepE &$63.3$&$61.9$&$62.6$&$60.8$&$67.5$&$64.3$&$65.5$&$63.5$\\
        SAPLMA & $\mathbf{90.2}$& $\mathbf{83.7}$&$69.7$&$67.8$&$69.3$&$64.5$&$75.6$&$69.5$\\
        \midrule
        SCGPT-PT & $\mathtt{N/A}$ & $\mathtt{N/A}$ &$72.3$&$70.3$&$76.4$&$69.9$&$72.5$&$72.5$\\
        SCGPT-NLI & $\mathtt{N/A}$ & $\mathtt{N/A}$ &$77.4$&$71.0$&$76.9$&$72.1$&$77.4$&$72.3$\\
        \midrule
        PPL-AVE &$67.1$&$64.1$&$61.8$&$59.2$&$61.7$&$58.5$&$63.7$&$63.3$\\
        PPL-MAX &$60.7$&$61.7$&$54.2$&$55.0$&$57.6$&$57.0$&$56.4$&$60.0$\\
        It-is-True &$54.8$&$60.1$&$63.0$&$59.5$&$61.7$&$59.1$&$75.6$&$71.0$\\
        \midrule
        \model &$86.6$&$82.1$& $\mathbf{80.4}$& $\mathbf{72.5}$& $\mathbf{83.9}$& $\mathbf{73.8}$& $\mathbf{83.3}$& $\mathbf{76.0}$\\
        \bottomrule
    \end{tabular}
    }
    \caption{Overall performance over four test sets.}
    \label{tab:overall}
\end{table}

\subsection{Main Results}
Table~\ref{tab:overall} presents the AUC and ACC scores for all the compared methods across four test sets. Meanwhile, Table~\ref{tb:efficiency} provides insights into the average detection time required by SCGPT-NLI and our \smodel for each instance. \textbf{In general, \smodel outperforms probing-based methods across all QA variation dataset, despite being trained without annotated labels. Additionally, \smodel exhibits superior performance compared to consistency checking methods and is also more efficient.} Some detailed findings include:

\textbf{Leveraging factuality labels substantially improves factuality detection accuracy.} 
This is evidenced by \smodel's superior performance trained on factuality labels, compared to confidence-based methods such as PPL and It-is-True. 

\textbf{Limitations of annotated labels on model transferability.}
Despite being trained using annotated labels, probing-based methods like RepE and SAPLMA lag behind a large margin compared to \smodel on the three QA variation test sets. This disparity arises because the two baselines are trained on True-False's training data, which consists of statements rather than question and responses. 
This difference in input distribution significantly limits the transferability of these models to out-of-distribution datasets. In contrast, \smodel is trained on a diverse range of questions, leading to superior performance across the QA datasets. The slight lag observed in True-False from SAPLMA is also attributed to the model's training on questions. However, probing the factuality of responses to questions is more practical than evaluating statements given by True-False. Therefore, training on datasets guided questions is reasonable.

\textbf{Self-consistency correlates well with factuality.} Despite SCGPT lacking supervision from factuality labels, it surpasses supervised probing-based baselines, suggesting a strong correlation between its self-consistency principle and factuality. Furthermore, \model, also adhering to the self-consistency principle, outperforms SCGPT. This is due to \smodel being exposed to numerous instances with diverse inconsistencies between responses, unlike SCGPT, which focuses solely on responses related to the given question. Moreover, \smodel evaluates the consistency of internal representations rather than discrete output responses like SCGPT, allowing it access to a wider range of information, thereby enhancing its predictive accuracy.

\textbf{\model's detection time is significantly shorter than SCGPT}, as shown in Table~\ref{tb:efficiency}. This is because \smodel relies on offline consistency checking, incorporating consistency characteristics into internal representations during training. As a result, its online inference depends solely on internal representations, eliminating the need for multiple online inferences like those performed by SCGPT.

\begin{table}[!t]
    \centering
    \scalebox{0.85}{
    \begin{tabular}{lcccc}
        \toprule
         &  NQ & TriviaQA & WebQ \\
        \midrule
        SCGPT-NLI &  2.530 & 2.210 & 2.050 \\
        \model & 0.024 & 0.023 & 0.024 \\
        
        \bottomrule
    \end{tabular}
    }
    \caption{Average time required for detecting each instance (in seconds).}

    \label{tb:efficiency}
\end{table}

\subsection{Cross-model Evaluation}
To implement \model, LLMs are invoked multiple times during the construction process, including data preparation, peer reviewing in consistency checking, and finally non-factual detection. 
By default, we employ the same LLM for all stages, leveraging an LLM's calibration property. 
Additionally, we explore whether training data generated by one LLM can effectively train a probe to detect the factuality of content generated by other LLMs. 
To study this, we use Llama2-7B for data preparation but vary the detection target to Llama2-13B and Mistral-7B. We further switch the LLM for consistency checking to Llama2-13B  and Mistral-7B, consistent with the model to be detected. It's important to note that for a given LLM to be detected, the probe needs to be aligned with it, specifically in terms of probe's input, which consists of the internal representation of the response along with the question, that must be generated by the LLM. The crucial findings, as presented in Table~\ref{tb:crossmodel}, include:

\textbf{More powerful LLMs brings better detection performance.} 
Comparing group 2 and 3 with group 1, where the training data remains consistent (created by Llama2-7B), probes built based on more powerful LLMs demonstrate higher performance, attributed to the enhanced representational capacity of these models.

\textbf{Generated data facilitates probing across various LLMs.} Switching the LLM for consistency checking to match the LLM being detected results in comparable performances between groups 4 and 2, as well as between groups 5 and 3, respectively.
This indicates that we can generate the training data, comprising (question, response, factuality label) triplets, once, regardless of the LLMs being probed, and utilize them uniformly to train probes for any LLM.

\begin{table}[t]
    \centering
    \scalebox{0.82}{
    \setlength{\tabcolsep}{5pt}
    \begin{tabular}{@{}c|ccccc@{}}
        \toprule
         \# & \tabincell{c}{Data\\Preparation} & \tabincell{c}{Consistency\\ Checking} & \tabincell{c}{Factuality\\detection} & AUC & ACC  \\
        \midrule
        1&Llama2-7B & Llama2-7B & Llama2-7B  & $80.4$ & $72.5$\\
        \midrule
        2&Llama2-7B & Llama2-7B & Llama2-13B & $81.1$ & $73.2$\\
        3&Llama2-7B & Llama2-7B & Mistral-7B & $81.3$ & $73.1$\\
        \midrule
        4&Llama2-7B & Llama2-13B & Llama2-13B & $81.7$ & $73.5$\\
        5&Llama2-7B & Mistral-7B & Mistral-7B & $81.4$ & $73.3$\\

        \bottomrule
    \end{tabular}
    }
\caption{Cross-model evaluation performance on NQ. We explore different combinations of LLMs across each of the three stages to evaluate their effectiveness.}

    \label{tb:crossmodel}
\end{table}

\subsection{Ablation Studies on Data Preparation}
We examine the impact of question distribution in the data preparation stage with two variants:

\begin{itemize}[leftmargin=*,itemsep=0pt,parsep=0pt]
    \item \textbf{\smodel with self questions:} 
    Assumes a scenario where the training dataset consists of questions from the same distribution as the test questions. 
    We utilize questions from the training data that belong to the same dataset as the test set.
    \item \textbf{\smodel with external questions:} Considers a scenario where questions from the same distribution as the test questions are unavailable. We utilize questions from the training data that belong to a different dataset from the test set.
\end{itemize}

\begin{figure*}[t]
	\centering
	\subfigure[NQ]{\label{subfig:nq}
		\includegraphics[width=0.21\textwidth]{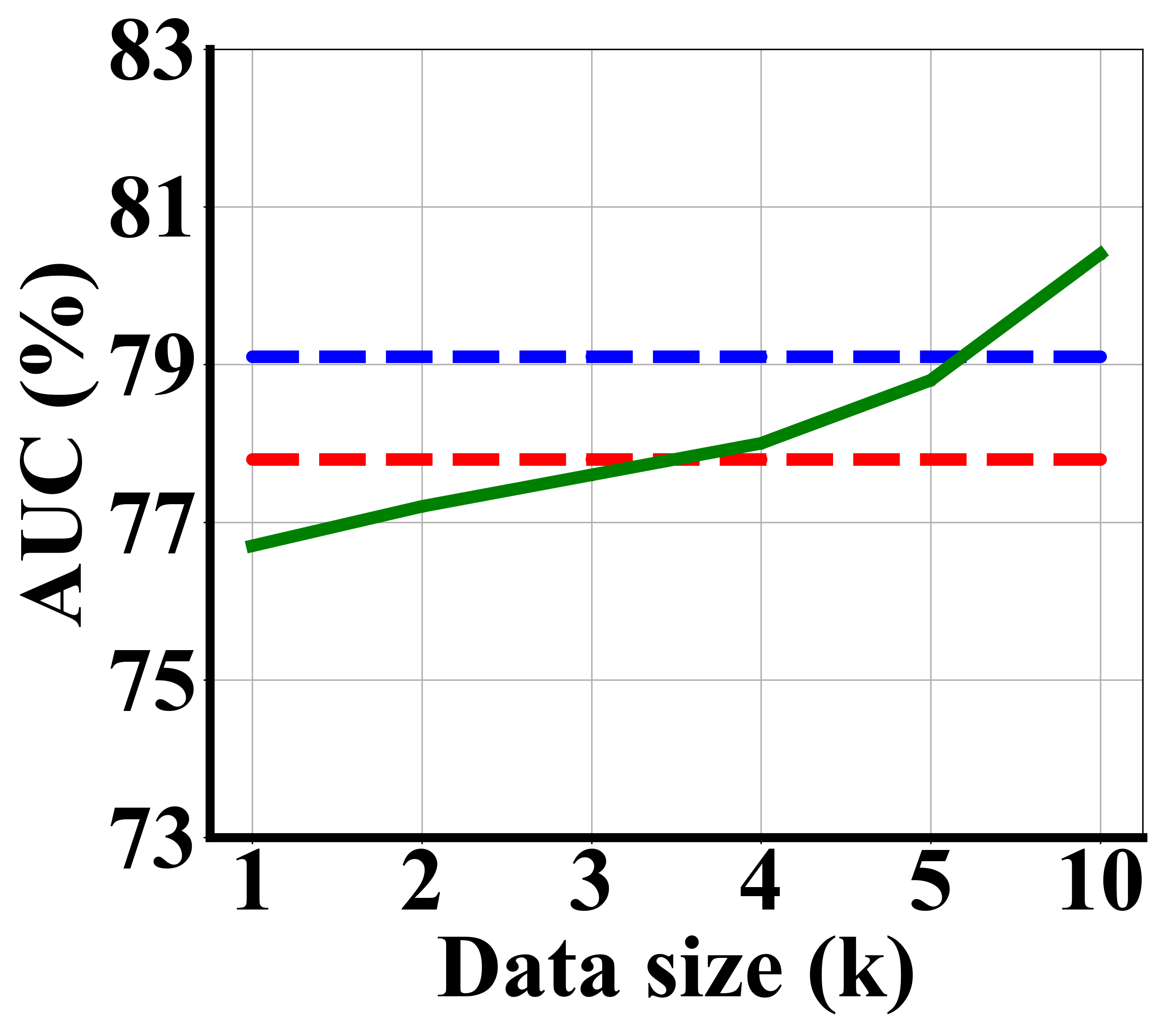}
	}		
	\subfigure[TriviaQA]{\label{subfig:triviaQA}
		\includegraphics[width=0.21\textwidth]{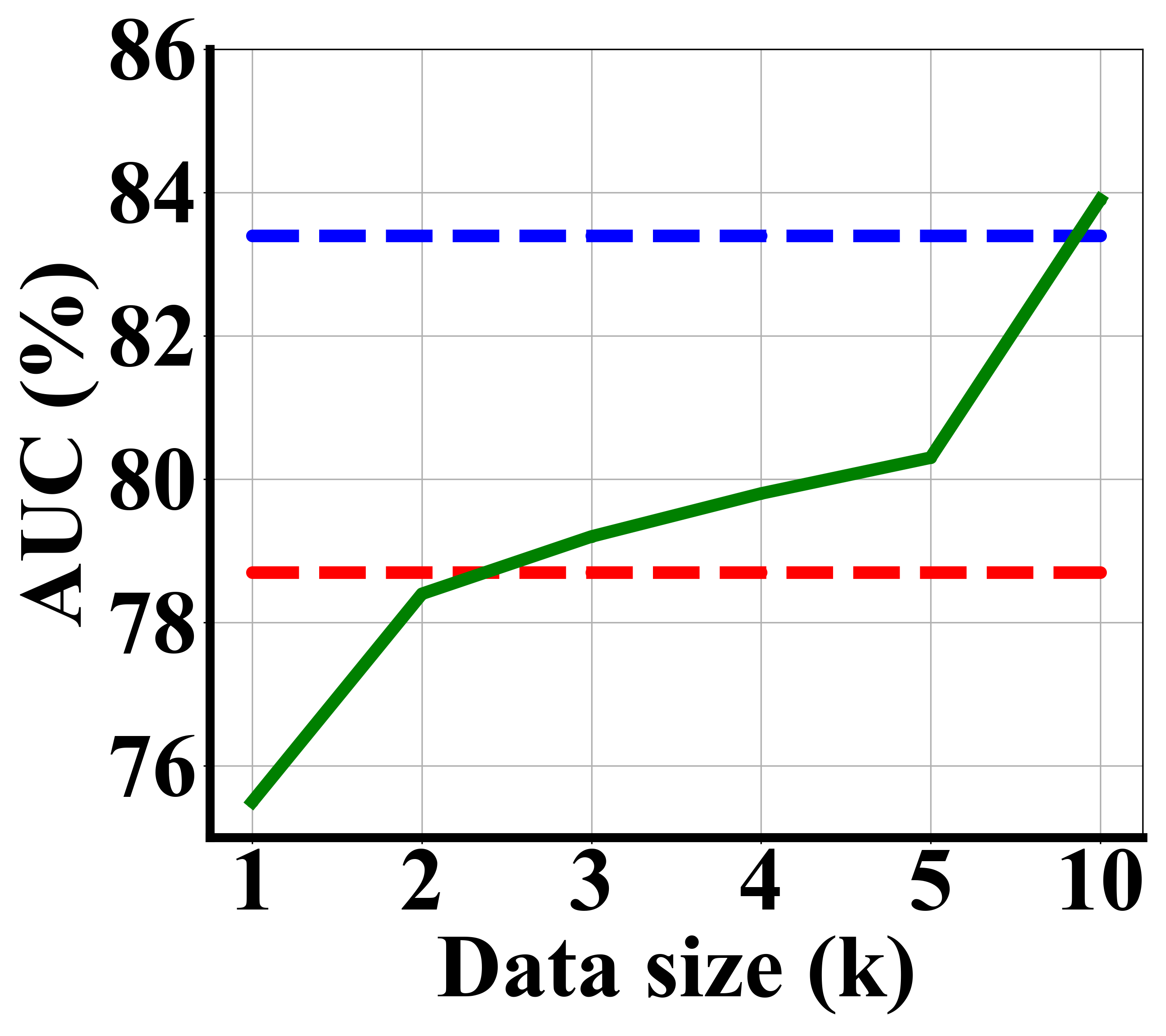}
	}		
	\subfigure[WebQA]{\label{subfig:webq}
		\includegraphics[width=0.21\textwidth]{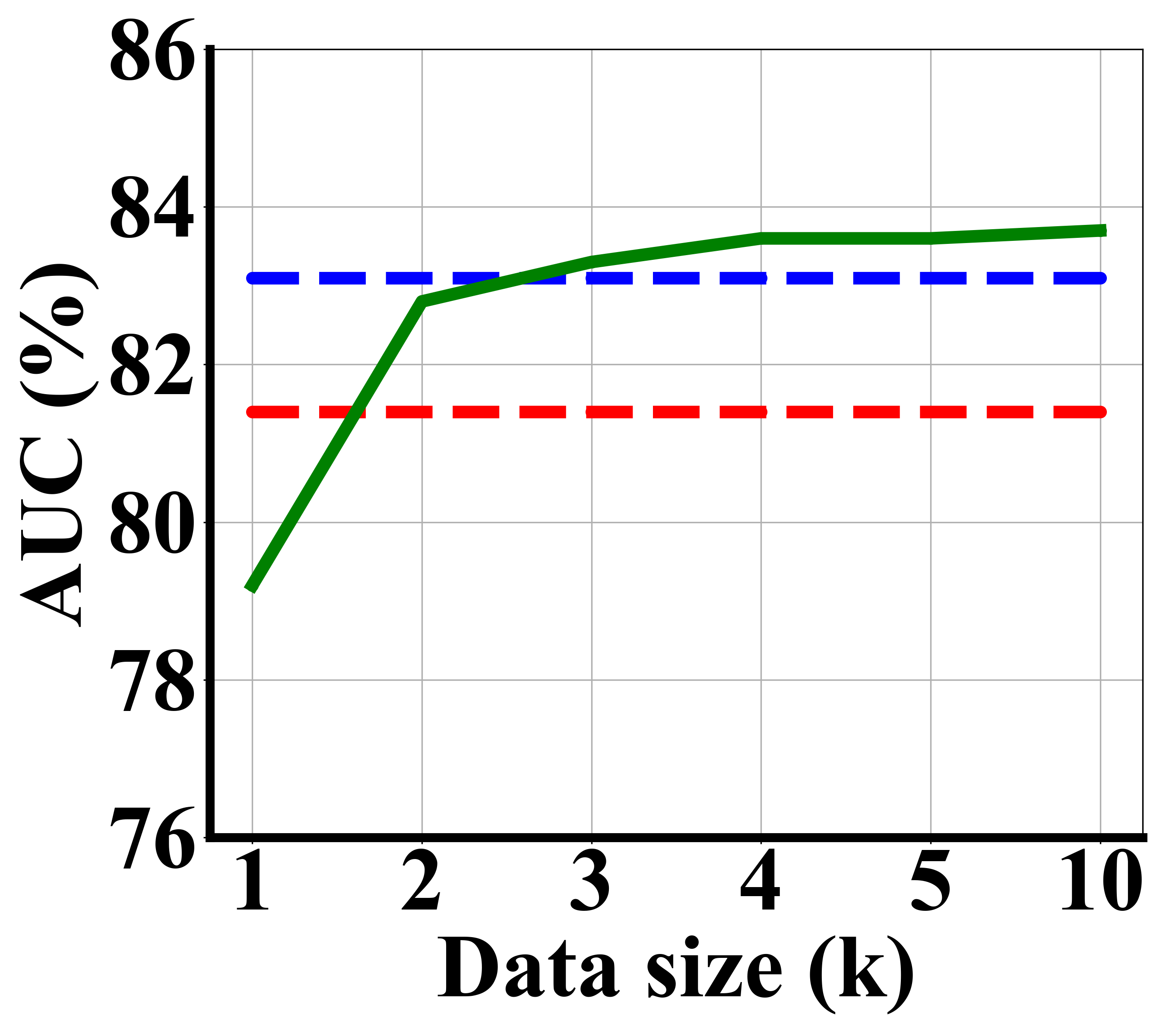}
	}		
	\subfigure[N\&k on NQ]{\label{subfig:consistencychecking}
		\includegraphics[width=0.21\textwidth]{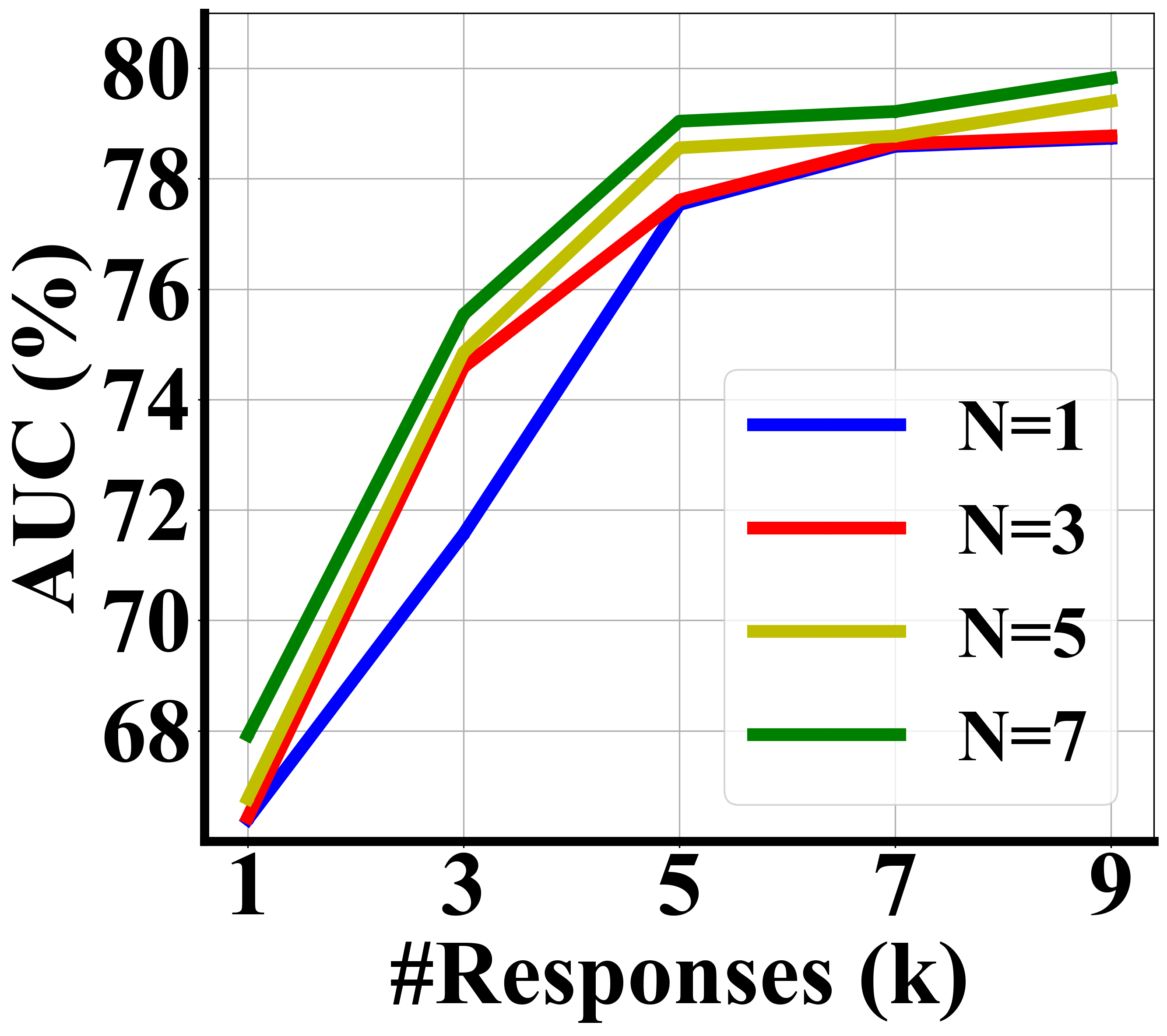}
	}		

 \caption{\label{fig:questiongeneration} Effects of question generation and the number of reviews and responses. We assess three question distributions for factual detection training data: ``\textcolor{blue}{\textbf{self questions}}'' ($1,000$ questions from the training data within the same dataset), ``\textcolor{red}{\textbf{external questions}}'' ($5,000$ questions from a different dataset), and our proposed approach, ``\textcolor{darkgreen}{\textbf{generated questions}}'' (without relying on available questions). Subfigures (a)-(c) demonstrate the effects of different question distributions on various test sets, while subfigure (d) presents the effects of various $k$ (the number of responses) and $N$ (the round of reviews per response) on NQ.}
\end{figure*}

We evaluate these variants across three QA test sets, maintaining a consistent number of training questions ($1,000$ per set) for the first variant. For the second variant, we collect $5,000$ training questions from the remaining two datasets for each target test set. Additionally, we vary the number of generated questions within the range of [1K, 2K, 3K, 4K, 5K, 10K] to assess its impact.
Figure~\ref{fig:questiongeneration}(a)(b)(c) presents the performance of these variants on the three QA test sets. We find:

\textbf{Training on questions of the same distribution as the test set yields significantly better results than a different distribution.} Despite having more external questions ($5,000$ vs $1,000$), the second variant still lags behind the first. 

\textbf{Training on generated questions could enhance transferability of the probe.} Across the three test sets, training with fewer than $5,000$ generated questions (approximately $3,000+$, $2,000+$, $1,000+$ questions on NQ, TriviaQA, and WebQA, respectively) can achieve performance comparable to using $5,000$ external questions. Additionally, training with approximately $6,000$, $10,000$, and $3,000$ generated questions on these three datasets respectively could outperform training using $1,000$ self-questions.
These results indicate that generated questions offer greater diversity, facilitating the transferability of the probe across different test sets, despite the diverse distributions observed among the three test sets.

\subsection{Ablation Studies on Consistency Checking}
We investigate the impact of two hyperparameters, $k$ (the number of responses) and $N$ (the number of review rounds per response), in the consistency checking stage.

Figure~\ref{fig:questiongeneration}(d) displays the detection performance on NQ with varying values of $k$ from 1 to 9 with interval 2 and different values of $N$ (1, 3, 5, 7). It's worth noting that $N=1$ corresponds to the review strategy in SCGPT.
Remarkably, the best performance is achieved with $N=7$, indicating that multiple inferences from an LLM, each guided by different demonstrations acting as instructions, contribute to more robust and confident review outcomes akin to opinions from multiple reviewers. The figure also illustrates that the performance exhibits a smooth increase as more responses are used, also suggesting that multiple responses could result in more confident consistency checking.

\subsection{Ablation Studies on Probe Construction}

We investigate feature selection at the probe construction stage, exploring the use of internal representations from the last (32nd), middle(16th), and first layers of Llama2-7B. Additionally, we experiment with averaging representations of all tokens within a layer or using only the last token. The default configuration includes the middle-layer representation and the last token in a layer.
The results, as depicted in Figure~\ref{tb:classifier}, indicate that the middle-layer representation and the last token are optimal choices within our setting.

\begin{table}[!t]
    \centering
    \scalebox{0.93}{
    \begin{tabular}{lcc}
        \toprule
         Model & AUC & ACC \\
        \midrule
        first-layer & $52.0$  & $53.3$\\
        middle-layer & $80.4$  & $72.5$\\
        last-layer & $76.4$  & $70.3$\\
        \midrule
        average token & $77.1$  & $70.9$\\
        last token & $80.4$  & $72.5$\\
        \bottomrule
    \end{tabular}
    }
    \caption{Evaluating probe construction using internal representations from different layers, either averaged or using the last token.}

    \label{tb:classifier}
\end{table}

\section{Related work}
\label{sec:related}

Factuality detection for LLM generated content mainly falls into two categories: consistency-based and probing-based.

Consistency-based methods detect non-factual generations by comparing model generated content with other information.
Among these methods, the most widely adopted assumption is that, LLMs usually fail to give consistent responses to the same prompt when generating multiple times~\cite{elazar2021measuring,mundler2023self,pacchiardi2023catch}, thus motivating a series method to detect non-factual content~\cite{selfcheckgpt,cohen2023lm,azaria2023internal} or reduce non-factual generations~\cite{dhuliawala2023chain,kuhn2023semantic}.
Another thread of works evaluate model self-consistency via model's confidence to the generated content~\cite{kadavath2022language,azaria2023internal,zou2023representation}.
Except for self-consistency checking, there are also attempts that use consistency between model generated content and external information as factuality indicator~\cite{wang2023factcheck,gao2023rarr,chern2023factool}.

Probing-based methods possesses the belief that the hidden representation entails certain property of generated content and can be extracted via a light weight model~\cite{alain2016understanding,gurnee2023language}.
Probing whether LLMs are producing factual content is proved to be feasible~\cite{kadavath2022language}, thus motivating researchers to develop more accurate probes~\cite{azaria2023internal,zou2023representation,chen2023hallucination}.
Comparing with these works, which rely on annotated training data, \model provides a method that distill consistency patterns from LLMs into a probe.

\section{Conclusion}
\label{sec:conclude}

This paper presents \model, a probing method for non-factual content detection that learns from offline consistency checking.
\model achieves good transferability among different distributed datasets as its does not rely on manually annotated data.
It also avoids the computational burden for online consistency checking.
In the future, \model potentially paves way to build more faithful LLMs.

\section*{Limitations}
The limitations of this work are as follows: 
(1) \textbf{Data Preparation:} \smodel employs an offline consistency checking method to automatically generate factuality labels for training a probe model. Although the probe model efficiently infers the factuality label of a response from an LLM in a single pass, the offline data preparation stage requires a large amount of data. This involves multiple inferences of the LLM for generating questions, responses, and reviews, resulting in high offline construction costs. Fortunately, as online usage increases, the amortized cost of offline construction decreases.
(2) \textbf{Open-sourced LLMs:} \smodel is limited to detecting factuality errors in open-sourced LLMs because it requires the internal representation of the input response along with the question as features input to the probing model for detection.
(3) \textbf{Factuality Error:} \smodel is constrained to detecting the factuality errosr in responses to questions or statements. Other aspects of errors, such as logical error, require further investigation.
\section*{Ethical Considerations}

We discuss the ethical considerations and broader impact of this work in this section:
(1) \textbf{Intellectual Property:} The datasets employed in this study, comprising True-False, NQ, TriviaQA, and WebQ, are widely accessible and established resources designed to facilitate extensive research in artificial intelligence and natural language processing (NLP). We are confident that these resources have been adequately de-identified and anonymized.
(2) \textbf{Data Annotation:} We recruit 10 annotators from commercial data production companies to label factual accuracy for three QA test sets, seed questions for question generation, and consistent/inconsistent response pairs for consistency checking. Annotators are compensated fairly based on agreed-upon working hours and rates. Prior to annotation, annotators are briefed on the data's processing and usage, which is formalized in the data production contract.
(3) \textbf{Intended Use:} The proposed \smodel is utilized for detecting non-factual content generated by LLMs.
(4) \textbf{Misuse Risks:} There is a risk that \smodel could be exploited for adversarial learning, potentially enabling LLMs to generate more implicit non-factual content that is more challenging to detect.
(5) \textbf{Potential Risk Control:} The trained \smodel is made publicly available to the open-source community, which may help mitigate the risks associated with its potential misuse for adversarial learning.
\bibliography{2-finalbib}

\appendix
\section{Appendix}
\startcontents[appendix]
\printcontents[appendix]{ }{1}{\setcounter{tocdepth}{2}\vspace{5pt}}

\subsection{Perplexity-based Baseline}
The perplexity-based baseline is formulated as:

\begin{align}
\text{PPL(AVE)} &= -\frac{1}{J}\sum_j \log p_{ij}, \\
\text{PPL(MAX)} &= \max_j (-\log p_{ij}),
\end{align}

\noindent where $i$ denotes the $i$-th statement, and $j$ denotes the $j$-th token in the $i$-th statement. $J$ is the number of tokens in the $i$-th statement. $p_{ij}$ denotes the probability of the $j$-th token in the $i$-th statement generated by the LLM. PPL(AVE) measures the average likelihood of all tokens, while PPL(MAX) measures the likelihood of the least likely token in the statement.

\begin{table}[htb]
    \centering
    \scalebox{0.8}{
    \begin{tabular}{@{}c|ccccc@{}}
        \toprule
         \# & \tabincell{c}{Data\\Preparation} & \tabincell{c}{Consistency\\ Checking} & \tabincell{c}{Factuality\\detection} & AUC & ACC  \\
        \midrule
        1&Llama2-7B & Llama2-7B & Llama2-7B  & 83.9 & 73.8\\
        \midrule
       2&Llama2-7B & Llama2-7B & Llama2-13B & 83.8 & 77.0\\
        3&Llama2-7B & Llama2-7B & mistral-7B & 86.8 & 76.8\\
        \midrule
        4&Llama2-7B & Llama2-13B & Llama2-13B & 83.7 & 77.1\\
        5&Llama2-7B & mistral-7B & mistral-7B & 86.3 & 76.9\\

        \bottomrule
    \end{tabular}
    }
\caption{Cross-model evaluation performance on TriviaQA.}
    \label{tb:crossmodel-TriviaQA}
\end{table}

\begin{table}[htb]
    \centering
    \scalebox{0.8}{
    \begin{tabular}{@{}c|ccccc@{}}
        \toprule
         \# & \tabincell{c}{Data\\Preparation} & \tabincell{c}{Consistency\\ Checking} & \tabincell{c}{Factuality\\detection} & AUC & ACC  \\
        \midrule
        1&Llama2-7B & Llama2-7B & Llama2-7B  & 83.3 & 76.0\\
        \midrule
        2&Llama2-7B & Llama2-7B & Llama2-13B & 83.4 & 76.3\\
        3&Llama2-7B & Llama2-7B & mistral-7B & 83.1 & 76.5\\
        \midrule
        4&Llama2-7B & Llama2-13B & Llama2-13B & 83.4 & 76.4\\
        5&Llama2-7B & mistral-7B & mistral-7B & 83.1 & 76.6\\

        \bottomrule
    \end{tabular}
    }
\caption{Cross-model evaluation performance on WebQ.}
    \label{tb:crossmodel-WebQ}
\end{table}

\begin{figure}[htb]
    \centering
    \includegraphics[width=0.30\textwidth]{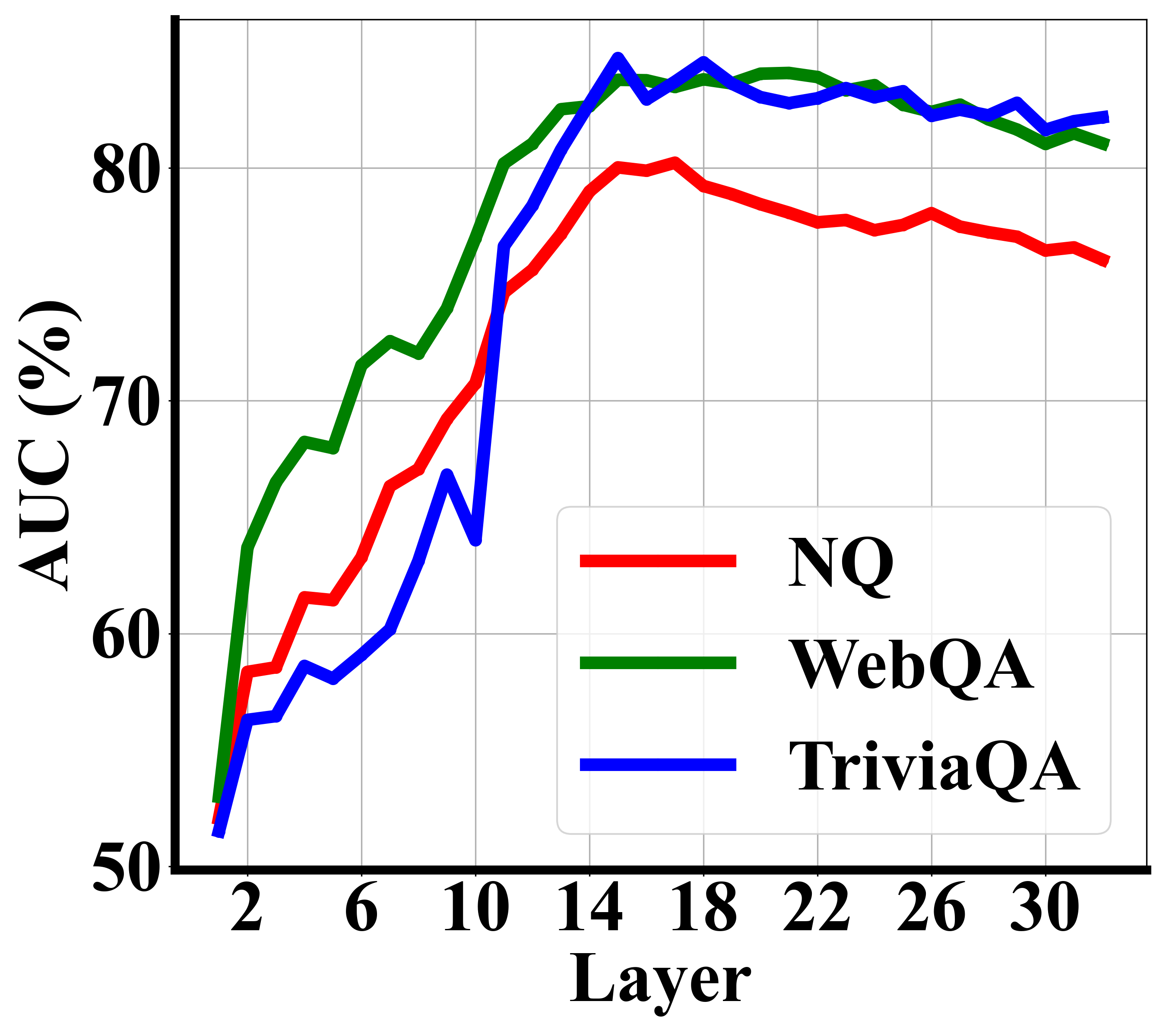}
    \caption{AUC obtained using the internal representations of different layers at the probe construction stage.}
    \label{fig:auc_at_layer}
\end{figure}

\subsection{Cross-model Evaluation on TriviaQA and WebQ}

We present the cross-model evaluation results on TriviaQA in Table~\ref{tb:crossmodel-TriviaQA} and on WebQ in Table~\ref{tb:crossmodel-WebQ} to explore whether training data generated by one LLM can effectively train a probe to detect the factuality of other LLMs. The settings are the same as those presented for NQ in Table~\ref{tb:crossmodel}. In addition to the default setting where we employ the same LLM for all stages of data preparation, consistency checking, and factuality detection, we also investigate two other settings. The first setting involves using Llama2-7B for data preparation but varying the detection target to Llama2-13B and Mistral-7B. The second setting involves further switching the LLM for consistency checking to Llama2-13B and Mistral-7B, consistent with the LLM to be detected. 

Tables~\ref{tb:crossmodel-TriviaQA} and \ref{tb:crossmodel-WebQ} present observations: for the first setting, when the detection target is changed to more powerful LLMs, the detection performance increases, indicating that more powerful LLMs can improve detection performance. Additionally, for the second setting, when changing the LLM for consistency checking to the same more powerful LLMs, the detection performance remains almost unchanged. This suggests that we can generate training data for questions, responses, and reviews once, regardless of the LLMs being probed, and uniformly employ them to train probes for any LLMs.

\subsection{Evaluation of Different Layers}
We vary the internal representations obtained per layer from the 1st to the last layer (32nd) of Llama2-7B, use them to construct probes respectively, and show the evaluated AUC on the three QA test sets in Figure~\ref{fig:auc_at_layer}. The results demonstrate that the detection performance of \model generally increases and then decreases with the increase in the number of layers. Typically, the best performance is achieved at the middle layer.

\subsection{Employed Prompts}
\label{sec:prompt}
We introduce the prompts used at different steps of our proposed \model, and also other baseline methods that need prompts. 

\vpara{Prompt for Question Generation in \model} is shown in Figure~\ref{prompt:questiongeneration}, where the seed questions are randomly sampled from an initial set of questions annotated by humans, which is then expanded by the newly generated questions.

\begin{figure*}
    \centering
    \begin{mdframed} [style=exampledefault,frametitle={Prompt for Question Generation in \model.}]
    \small
Please ask some objective questions of similar difficulty to [Seed Questions].\\\\
\begin{color}{blue}\\
\#\#\# [Seed Questions]\\
1. which part of earth is covered with water?\\
2. what is the military equivalent of a gs-14?\\
3. who provided the voice for the geico insurance company gecko?\\
4. who played the father in sound of music?\\
5. fugees killing me softly with his song original?\\
6.
\end{color}\\
    \vspace{1em}
    \end{mdframed}
    \caption{Prompt for question generation in \model. Five seed questions are provided and the blank following item 6 is the new question that encourages LLMs to generate.}
    \label{prompt:questiongeneration}
\end{figure*}

\vpara{Prompt for Response Generation in \model} is shown in Figure~\ref{prompt:responsegeneration}, where five instructions for generating responses are presented. We randomly select an instruction from this set each time to encourage diverse response generation.

\begin{figure*}
    \centering
    \begin{mdframed} [style=exampledefault,frametitle={Prompt 1 for Response Generation in \model.}]
    \small
\#\#\# Question\\
where is taurus the bull in the night sky\\
\#\#\# Answer\\
    \vspace{1em}
    \end{mdframed}
    \begin{mdframed} [style=exampledefault,frametitle={Prompt 2 for Response Generation in \model.}]
    \small
\#\#\# Instruction\\
\begin{color}{blue}
Answer the following question.\\\\
\end{color}\\
\#\#\# Question\\
where is taurus the bull in the night sky\\
\#\#\# Answer\\
    \vspace{1em}
    \end{mdframed}
    \begin{mdframed} [style=exampledefault,frametitle={Prompt 3 for Response Generation in \model.}]
    \small
\#\#\# Instruction\\
\begin{color}{blue}
Give a helpful answer.\\\\
\end{color}\\
\#\#\# Question\\
where is taurus the bull in the night sky\\
\#\#\# Answer\\
    \vspace{1em}
    \end{mdframed}
    \begin{mdframed} [style=exampledefault,frametitle={Prompt 4 for Response Generation in \model.}]
    \small
\#\#\# Instruction\\
\begin{color}{blue}
Generate a brief response in just one sentence.\\\\
\end{color}\\
\#\#\# Question\\
where is taurus the bull in the night sky\\
\#\#\# Answer\\
    \vspace{1em}
    \end{mdframed}
    \begin{mdframed} [style=exampledefault,frametitle={Prompt 5 for Response Generation in \model.}]
    \small
\#\#\# Instruction\\
\begin{color}{blue}
Compose a concise answer within a single sentence.\\\\
\end{color}\\
\#\#\# Question\\
where is taurus the bull in the night sky\\
\#\#\# Answer\\
    \vspace{1em}
    \end{mdframed}
    \caption{Prompt for response generation in \model. Five different instructions are randomly employed to elicit diverse responses.}
    \label{prompt:responsegeneration}
\end{figure*}

\vpara{Prompt for Consistency Checking in \model} is depicted in Figure~\ref{prompt:reviewgeneration}. It instructs LLMs to determine whether two responses are ``\texttt{Consistent}'', ``\texttt{Neutral}'', or ``\texttt{Non-Consistent}'' given a question. For each judgment, three demonstrations are randomly selected from a set of 16 consistency judgment pairs. These diverse demonstrations facilitate the collection of multiple judgments (reviews) for each comparison between two responses, potentially offering unique perspectives.

\begin{figure*}
    \centering
    \begin{mdframed} [style=exampledefault,frametitle={Prompt for Consistency Checking in \model.}]
    \small
Assess the connection between the two responses to the initial query, taking into account the potential scenarios of Endorsement, Contradiction, and Impartiality.\\

\begin{color}{blue}
\#\#\# Input\\\\
- **Question:** where is vina del mar located in chile \\
- **First Response:** Vina del Mar is located in the Valparaíso Region of Chile, approximately 120 kilometers west of Santiago. \\
- **Second Response:** Viña del Mar is a city located on the central coast of Chile. It is part of the Valparaíso Region and is situated about 120 kilometers (75 miles) northwest of Santiago, the capital of Chile. \\\\
\#\#\# Output\\
Judgement: Endorsement\\
@Reason@: The two responses provide consistent information about the location of Viña del Mar in the Valparaíso Region of Chile, approximately 120 kilometers west/northwest of Santiago. The details in both responses align, endorsing the accuracy of the information.\\\\

\#\#\# Input\\\\
- **Question:** where is taurus the bull in the night sky\\
- **First Response:** Taurus the Bull is located in the southeastern part of the sky, near the constellation Orion and the celestial equator.\\
- **Second Response:** Taurus the Bull is located in the eastern part of the night sky, stretching from the constellation Orion to the constellation Gemini.\\\\
\#\#\# Output\\
Judgement: Contradiction\\
@Reason@: The two responses provide different information regarding the location of Taurus the Bull in the night sky.\\\\

\#\#\# Input\\\\
- **Question:** who is the old man in waiting on a woman\\
- **First Response:** The old man in the waiting room is Mr. Johnson's father.\\
- **Second Response:** The old man in the picture is likely the grandfather or great-grandfather of the woman he is waiting on, as he appears to be elderly and has a gentle expression on his face.\\\\
\#\#\# Output\\
Judgement: Impartiality\\
@Reason@: There is no explicit contradiction between the two responses, and they may collectively provide a more detailed and comprehensive answer to the question. The overall tone is impartial, as the information in the responses is neither conflicting nor mutually supportive.\\\\
\end{color}

\#\#\# Input\\\\
- **Question:** 
\begin{color}{blue}
where is vina del mar located in chile\\
\end{color}
- **First Response:** \begin{color}{blue}
Vina del Mar is located in the Valparaíso Region of Chile, approximately 120 kilometers west of Santiago.\\
\end{color}
- **Second Response:** \begin{color}{blue}
Viña del Mar is a city located on the central coast of Chile. It is part of the Valparaíso Region and is situated about 120 kilometers (75 miles) northwest of Santiago, the capital of Chile.
\end{color}\\\\
\#\#\# Output\\
Judgement:
    \vspace{1em}
    \end{mdframed}
    \caption{Prompt for Consistency Checking in \model. Three demonstrations illustrate judgments of endorsement, contradiction, and impartiality, respectively.}
    \label{prompt:reviewgeneration}
\end{figure*}

\vpara{Prompt for Probe Construction in \model} is depicted in Figure~\ref{prompt:probeconstruction}, where a question and the answer generated by the LLM under detection are organized as input for the LLM to obtain their internal representation. This representation then serves as input for the probe to predict its factual label.

\begin{figure*}
    \centering
    \begin{mdframed} [style=exampledefault,frametitle={Prompt for Probe Construction in \model.}]
    \small
\#\#\# Instruction\\
Compose a concise answer within a single sentence.\\\\
\#\#\# Question\\
\begin{color}{blue}
where is taurus the bull in the night sky\\
\end{color}\\
\#\#\# Answer\\
\begin{color}{blue}
Taurus the Bull is located in the southeastern part of the sky, near the constellation Orion and the celestial equator.\\
\end{color}\\
    \vspace{1em}
    \end{mdframed}
    \caption{Prompt for Probe Construction in \model. The LLM under detection receives the question and its generated answer to obtain the corresponding internal representation. This representation serves as the input for training the probe.}
    \label{prompt:probeconstruction}
\end{figure*}

\vpara{Prompt for SelfCheckGPT-Prompt} is displayed in Figure~\ref{prompt:selfcheckGPT}, identical to the one presented in \cite{selfcheckgpt}. SelfCheckGPT-Prompt facilitates consistency checking by presenting a sentence and its context to LLM, enabling it to judge whether the context adequately supports the sentence.

\begin{figure*}
    \centering
    \begin{mdframed} [style=exampledefault,frametitle={Prompt for SelfCheckGPT with Prompt.}]
    \small
Context: \begin{color}{blue}
Delhi was made the capital of India for the first time by the British East India Company in 1858, when the British assume control of the Indian subcontinent following the Indian Rebellion of 1857.\\
\end{color}\\\\
Sentence: \begin{color}{blue}
Delhi was first made the capital of India by the Mughal emperor Shah Jahan in the 17th century.\\
\end{color}\\\\
Is the sentence supported by the context above? Answer Yes or No.\\\\
Answer: 
    \vspace{1em}
    \end{mdframed}
    \caption{Prompt for SelfCheckGPT-Prompt. A sentence and its context are provided to enable LLMs to determine whether the sentence is supported by the context.}
        \label{prompt:selfcheckGPT}
\end{figure*}

\label{sec:appendix}


\end{document}